\documentclass[runningheads]{llncs}
\usepackage{graphicx}
\usepackage{multirow}
\usepackage{makecell}
\usepackage{amssymb}
\usepackage[ruled,vlined]{algorithm2e}
\usepackage{subcaption}
\usepackage{url}

\begin{document}
\title{Data Fusion for Deep Learning on Transport Mode Detection: A Case Study}
%
%
\author{Hugues Moreau\inst{1, 2}\orcidID{0000-0002-0569-4190} \and
Andrea Vassilev\inst{1} \and
Liming Chen\inst{2} }
\authorrunning{H. Moreau et al.}
%
\institute{Commissariat à l'Energie Atomique \email{\{hugues.moreau, andrea.vassilev\}@cea.fr } \and
Ecole Centrale de Lyon, \email{\{hugues.moreau,liming.chen\}@ec-lyon.fr}}
\maketitle              
\begin{abstract}
In Transport Mode Detection, a great diversity of methodologies exist according to the choice made on sensors, preprocessing, model used, etc. In this domain, the comparisons between each option are not always complete. Experiments on a public, real-life dataset are led here to evaluate carefully each of the choices that were made, with a specific emphasis on data fusion methods. Our most surprising finding is that none of the methods we implemented from the literature is better than a simple late fusion. Two important decisions are the choice of a sensor and the choice of a representation for the data: we found that using 2D convolutions on spectrograms with a logarithmic axis for the frequencies was better than 1-dimensional temporal representations.
To foster the research on deep learning with embedded inertial sensors, we release our code along with our publication.

\end{abstract}

This is a preprint of the following chapter: Moreau, H, Vassilev, A., Chen, L., Data Fusion for Deep Learning on Transport Mode Detection: A Case Study, published in Proceedings of the 22nd Engineering Applications of Neural Networks Conference, edited by Lazaros Iliadis, John Macintyren Chrisina Jayne, and Elias Pimenidis, 2021, Springer Nature Switzerland AG 2021, reproduced with permission of Springer Nature Switzerland AG 2021. The final authenticated version is available online at: \url{https://doi.org/10.1007/978-3-030-80568-5_12}.

\section{Introduction}

Transport mode detection is a classification problem aiming to design an algorithm that can infer the transport mode of a user given multimodal signals (GPS and/or inertial sensors). It has many applications, such as carbon footprint tracking, mobility behaviour analysis, or real-time door-to-door smart planning.
The signals are collected from an embedded device (either the sensors of a mobile phone, or a dedicated device), and processed by a Transport Mode Detection Algorithm, in order to know the transport mode of the owner of the device. This algorithm has multiple steps that often include a preprocessing step, and classification in itself. The classification step is where algorithms differ the most from each other. All algorithms use Machine Learning \textit{i.e.}, methods that use a certain amount of labeled data to learn how to predict the output transport mode, before making predictions on unseen samples. 

Contrary to domains like Computer Vision, the approaches in this domain differ greatly from each other: preprocessing, model, chosen sensors, recording conditions, etc. This diversity of variables further prevent fair and efficient comparisons between publications. Recently, the Sussex-Huawei Locomotion team published a dataset containing data from inertial sensors (accelerometer, gyrometer, magnetometer, etc.), in order to organize a  challenge: they published a certain amount of labeled and unlabeled data, and asked researchers to make predictions on the unlabeled data set. As they evaluated each prediction using unseen labels, they could establish clear comparisons between approaches. However, several many options typically differ between two publications, thus preventing comparisons. For instance, all the approaches \cite{choi_confidence-based_2018,gjoreski_applying_2018,ito_application_2018,widhalm_top_2018} had different methods, preprocessing pipelines, architectures, etc. even though they worked with on same exact problem. We start from one of them, and evaluate carefully each of the choices that were made, with a specific emphasis on data fusion methods.

Our main contributions are the following: 
\begin{itemize}
    \item We measure the performance of each sensor: the single best sensor is the norm (magnitude) of the acceleration, and the second best real sensor is the gyrometer  (section \ref{subsec:results_per_sensor}).
    \item We show the best preprocessing method is to resort to spectrograms (time-frequency diagrams), to compute the log of the power, with a log scale for the frequencies  (section \ref{subsec:results_preprocessing}). We go further and measure numerically the importance of each of the parameters. 
    \item We evaluate 13 different data fusion methods and show that they are, for the most part, equivalent. (section \ref{subsec:results_fusion_modes})
\end{itemize}

In a hope to help reproduce and foster research on this subject, we also publish the code that allowed us to generate the results.

The rest of the publication is organized as follows: section \ref{sec:related_works} reviews the publications that focus on Human Activity Recognition and data fusion. Sections \ref{sec:baseline} and \ref{sec:experiments_presentation} present the single-modality architecture, and the different options we will compare. Finally, we detail the results of the experiments and in section \ref{sec:experiments_results}.

\section{Related works}\label{sec:related_works}

\subsection{Transport Mode Detection}

Transport Mode Detection (TMD) is a  classification problem whose goal is to detect the transport mode of a human subject, \textit{e.g.}, walking, cycling, using any sensor. With inertial sensors, multiple methods coexist: some use handcrafted features with traditional learning methods (see \cite{gjoreski_applying_2018,janko_cross-location_2019,alotaibi_transportation_2020}, for instance), while others use directly use Deep Neural networks (Convolutional \cite{ito_application_2018} or Recurrent Neural Networks \cite{ordonez_deep_2016}). 
For inertial sensors, the most important dataset is the SHL dataset, which is used to organize a yearly challenge \cite{wang_summary_2018,wang_summary_2019,wang_summary_2020}. In the next section, we will present the dataset, the challenge that comes with it, and provide an outline of the best participations.

\subsubsection*{The SHL 2018 challenge}

The SHL 2018 challenge is a supervised classification problem. The participants were given a series of 16,310 consecutive annotated recordings of embedded sensors and had to classify some 5978 samples of a test set. Each recording is 60-seconds long, and contains data from 7 sensors: accelerometer, magnetometer, gravity, linear acceleration (acceleration minus gravity), gyrometer, orientation quaternion, and barometric pressure. Each signal was recorded at $100 Hz$, so that one sample to classify is a set of 20 vectors of size $60 \times 100 = 6000$ points. There are 8 classes available: Still, Walk, Run, Bike, Car, Bus, Train, Subway. Our protocol mimics the organisation of the challenge: we make our choices on a validation dataset, the annotations on the 5978 test samples are used only in section \ref{subsec:training_protocol}.

The most successful method is the submission from the Joseph-Stephan Institute \cite{gjoreski_applying_2018}. This approach used an ensemble of methods (both Machine Learning and Deep Learning), with a hidden markov model as meta-classifier to return a prediction. They reached a F1-score of 93.86 \%  on the unseen test set, ranking first on the 2018 challenge. 

The next best team \cite{ito_application_2018} used spectrograms (time-frequency diagrams), along with specific preprocessing. The 'images' containing the log of the power were then given to a 2-dimensional CNN for classification. This pure-deep learning approach is the one we selected as our baseline (section \ref{subsec:preprocessing}). But the participation only used two sensors: the accelerometer with gyrometer. For each classification, the spectrograms from these two sensors were concatenated along their 'frequency' axis, to form a single image which was to be classified by a single network. In the challenge, this reached a 88.83 \% F1-score. 

Several other participants did submit a prediction to the challenge. Similarly to the Transport Mode Detection literature, some used traditional ML algorithms with handcrafted features, others relied on Recurrent Neural Networks, or a combination of CNN and RNN. Wang \textit{et al.} \cite{wang_summary_2018} published a complete synthesis of the challenge participations along with the results. However, they did not conduct any experiments to assess the importance of each possible choice.

Since the end of the challenge, other publications worked on this dataset, sometimes using a different test set. The best one is from Gjoreski \textit{et al.} \cite{gjoreski_classical_2020}, who improved the model that scored first in 2018 \cite{gjoreski_applying_2018}. By adding another neural network to the prediction models, they managed to improve the F1-score on the SHL test set by one percentage point, up to 94.9\%. 

Most methods working with traditional Machine Learning models use a simple fusion: once all features are computed for all sensors, they are concatenated and given to a classifier (\cite{gjoreski_applying_2018,janko_cross-location_2019,alotaibi_transportation_2020}). With deep learning models, the data fusion methods change much more between publications. In the next section, we will present the diverse fusion modes used in Deep Learning approaches.

\subsection{Data Fusion modes in deep learning}

Most approaches rely on simple fusion modes: Early fusion (concatenation of input signals \cite{ito_application_2018,wang_deep_2019}), intermediate fusion (concatenation of representation coming from different sensors \cite{zeng_convolutional_2014}), or late fusion (average of predictions \cite{richoz_transportation_2020})

Some approaches are sensor-specific, either because they work on textual data \cite{ma_multimodal_2015} and rely on the specific structure of the medium; or because they create an explicit alignment between sensor data with different ranges (\textit{eg} a RGB camera looking forward and a LIDAR sensor gathering information from all directions \cite{feng_leveraging_2020}), which is not necessary in our case as the signals from different sensors are already synchronized. Others focus on pretraining by reproducing one modality with the other, in case one modality is suddenly missing \cite{ngiam_multimodal_2011}. 

But some methods still rely on relevant principles: for instance, Wang \textit{et al.} \cite{wang_what_2019} work with audiovisual videos, and create a model whose weights minimize the overfitting. Chen \textit{et al.} \cite{chen_selective_2019} designed a network that attributes a weight to each sensor, similarly to attention \cite{xu_show_2016}. Liu \textit{et al.} \cite{liu_learn_2018} tried to conceive a network that does not require every sensor to be good everytime. 

These methods are always better than the state of the art on the dataset each publication considered, but most work provide little comparisons with other fusion methods, if any. In general, most of the publications which deal with multiple sensors compare their architecture to a small subset of baselines, and those subsets do not always overlap between publications. Reviews exist to identify the different fusion modes \cite{gao_survey_2020,liu_urban_2020}, but they only report the performance each article provided. To our knowledge, none compare the performance of each method with the same exact sensors, architectures, and evaluation method. We provide a clear comparison of the different fusion modes, including the most basic ones.

\section{Baseline }\label{sec:baseline}

\subsection{A network using a single modality}\label{sec:baseline_archi}
Before considering the fusion of different sensor data, we need to consider a network that succeeds at classifying signals from a single sensor. We rely on the architecture in \cite{ito_application_2018}: the Convolutional Neural Network  has three convolutional layers, a flatten step, and two fully-connected layers. The hyperparameters and training process are the same as in \cite{ito_application_2018}. 

We will consider this network as our baseline to evaluate fusion methods: all fusion methods that use multiple sensors should at least be better than a network using only the best sensor (the accelerometer).

\subsection{Training protocol}\label{subsec:training_protocol}

The 2018 SHL Challenge asked candidates to give one prediction per timestamp (that is, to output 6,000 predictions per sample) but, as only 4\% of the samples of the database have more than one mode, we work on a simpler problem: each sample in the dataset is assigned a single transport mode, which is the mode at $t = 30s$. 
We split the train and validation splits similarly to \cite{widhalm_top_2018}: to avoid contamination between training and validation samples, we sort the data chronologically, and we send the first 3,000 samples to the validation set, while the last 13,000 samples go into the train set (we use the first samples for the validation set because the end of the dataset is severely imbalanced).

To generate the results of table \ref{tab:results_preprocessing} to \ref{tab:results_fusion_modes}, the network is trained for 50 epochs on the training set, before being evaluated on the validation set. Each evaluation is repeated 5 times (with a new random seed every time), the mean and standard deviation of the F1 score are given as a result. 
To test the best method against the state of the art, we train the model with the union of the train and validation set, and evaluate our results on the test set. The mean and standard deviation are given in section \ref{subsec:test_results}.

\section{Experiments}\label{sec:experiments_presentation}
We design three experiments where we evaluate a series of alternatives. As evaluating all the possible choice combinations is impossible, we evaluate each choice one by one. 
In most situations, when evaluating one parameter, we need to choose a value for the other parameters. In these cases, the other parameters are set to their \textit{baselines}, which we will present in each case.

\subsection{Sensor choice}
To confirm our choice of sensors, we decide to evaluate each all the sensors individually. For each sensor available (accelerometer, gyrometer, etc.) we consider every possible axis (\textit{x, y, z}, with the possible addition of \textit{w} for the orientation quaternion), in addition to the norm of these axes (computed using the euclidean norm). The norm is hoped to represent an orientation-independent version of the signals. We compute a log-power of the spectrograms with a log axis, and evaluate each signal individually.

\subsubsection*{baseline}
\begin{figure}
    \centering
    \includegraphics[width=12cm]{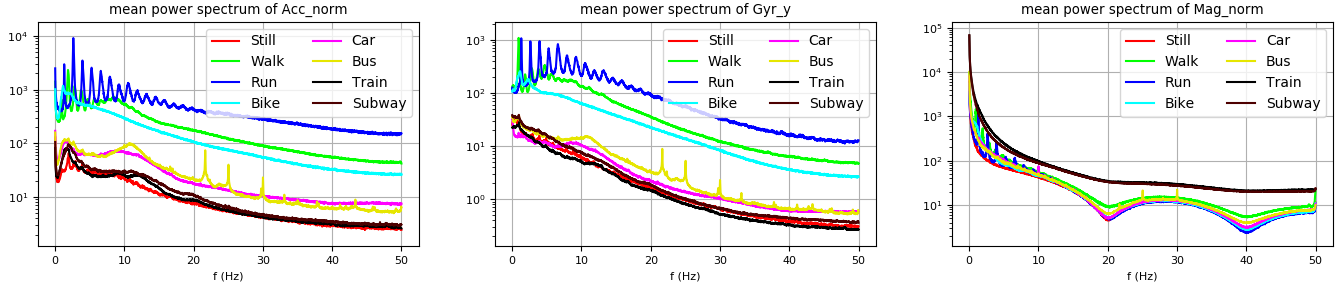}
    \caption{The average power spectrum per class (only half of the spectrum is shown). }
    \label{fig:avg_spectrum}
\end{figure}

The design of a heuristic is out of scope of this work. To make a choice, we simply selected the sensors using prior knowledge and related literature. We choose the norm of the accelerometer, for it is the single best signal available (\textit{cf} table \ref{tab:results_per_sensor}). As in \cite{ito_application_2018}, we add the $y$-axis of the gyrometer. But those two sensors only measure the inertial dynamics. To add a different kind of information, we choose to use the norm of the magnetometer. In exterior, this signal capture the Earth's magnetic field, which does not change much. It  varies greatly around metal objects and strong electrical currents. 
We expect this means the sensor to help to distinguish the Still class from the motorized ones (Train, Subway). This can be checked by looking at the average power spectrum of each class (Fig. \ref{fig:avg_spectrum}).

When considered alone, the norm of the magnetometer is worse than any of its three individual axes ($x, y, z$, table \ref{tab:results_per_sensor}). This is because the axes can act as a compass, thus retaining information about the dynamics of the movement. Computing the norm destroys this information, which is not a problem in our setting as we always use the magnetometer along with other inertial sensors (accelerometer and gyrometer). 
As a sanity check, we add the $w$ component of the orientation vector, which encodes the amount of rotation the phone detects.
We expect this signal to carry similar information to the gyrometer, which means adding it to the triplet ($|Acc|, Gyr_y, |Mag|$) should not improve results significantly.

\subsection{Preprocessing}\label{subsec:preprocessing}

For our comparisons, we consider the 1-dimensional temporal data, and different kinds of spectrograms. For each signal, we also compute the power spectrum using the Fast Fourier Transform (FFT). This is intended to be halfway between the 1D temporal and 2D spectrogram representations. For the results in the 'temporal' and 'FFT' categories (and for these results only), the network uses 1-dimensional convolutions, with filters of size $3$. All the other parameters remain similar to \cite{ito_application_2018}.  

\subsubsection*{baseline}
\begin{figure}[t]
\centering
    \begin{subfigure}[b]{0.4\textwidth}
         \centering
         \includegraphics[width=1.1in]{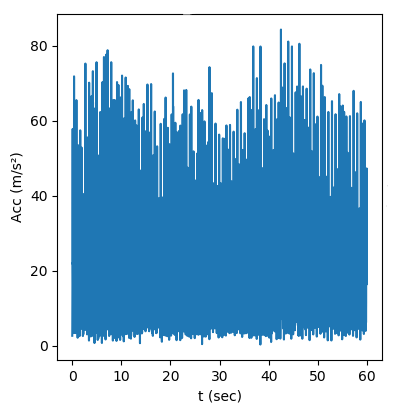}
         \caption{Raw data}
         \label{fig:preprocessing_outline:temporal}
    \end{subfigure}
    \begin{subfigure}[b]{0.4\textwidth}
         \centering
         \includegraphics[width=2in]{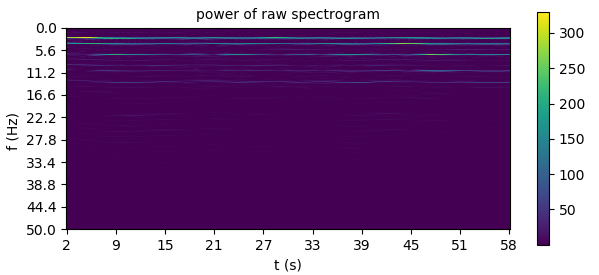}
         \label{fig:preprocessing_outline:spectro}
         \caption{full-size spectrogram}
    \end{subfigure}

    \begin{subfigure}[b]{0.4\textwidth}
         \centering
         \includegraphics[width=1.1in]{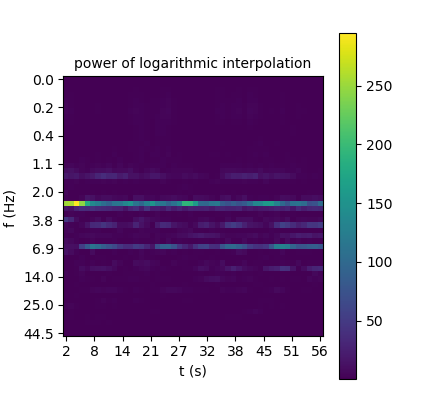}
         \caption{using a log for the frequency interpolation}
         \label{fig:preprocessing_outline:spectro_logfreq}
    \end{subfigure}
    \begin{subfigure}[b]{0.4\textwidth}
         \centering
         \includegraphics[width=1.1in]{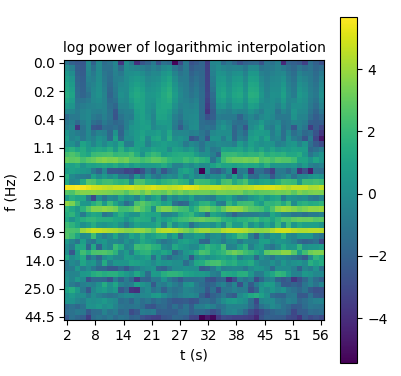}
         \caption{using a log for the power scale}
         \label{fig:preprocessing_outline:spectro_logfreq_logpow}
    \end{subfigure}
\caption{An illustration of the preprocessing step with the norm of the accelerometer data from a running segment. }
\label{fig:preprocessing_outline}
\end{figure}

We repeat the preprocessing protocol in \cite{ito_application_2018}. Each temporal signal is first converted into a spectrogram using a moving STFT window. The samples are 6,000 points long segments (60 seconds at 100 Hz), and we use 5 seconds-long windows with 4.9s overlap. Then, the spectrograms are rescaled into $48 \times 48$ pixels (this exact resolution was chosen to fit the architecture of the network, see section \ref{sec:baseline}). The time axis is rescaled linearly, while the frequency axis is rescaled using a logarithmic interpolation (similarly to the mel scale). This allows to give more importance to the lowest frequencies (Fig. \ref{fig:avg_spectrum} shows these frequencies are paramount). Lastly, we compute the log of the power, in order to make visible regions with different orders of magnitude (See fig. \ref{fig:preprocessing_outline} for an illustration).

\subsection{Fusion modes}\label{subsec:fusion_modes}
We selected several data fusion methods that are relevant in our setting for comparison. The names may vary, some methods might even not be named. 
One important note: most of these modes are equivalent to our baseline architecture (section \ref{sec:baseline}) when a single modality is used (if they leave the possibility to use a single sensor). The exceptions are the bottleneck filters and attention.

\begin{itemize}
    \item \textit{Early fusion} modes consist in giving all signals to a single neural network. The only difference between them is how the input signals are put together before they are processed by the neural network.
    \begin{itemize}
        \item \textit{Time concatenation}: the input spectrograms are concatenated along their temporal axis.
        \item \textit{Frequency concatenation}: the input spectrograms are concatenated along their frequency axis (after log interpolation).
        \item \textit{Depth concatenation}: the signals are put together like the channels of a RGB image: each convolution filter of the input layer has access to the same portion of all signals at the same time. 
    \end{itemize}
    Note that early fusion is not always possible for every problem. As Moya-Rueda \textit{et al.} notice \cite{moya_rueda_convolutional_2018} using these modes require to have the sensors synchronized (which implies having the same frequency) in order for the fusion to be relevant.

    \item \textit{Intermediate fusion} consists in merging together the features produced by different sensor-specific networks, so that a single classification network can process them. As it requires the classifier to have some kind of internal representation, it most adapted for deep architectures. 
    \textit{Feature concatenation}: The idea behind feature fusion is to let the convolutional layers compute features, before giving both features to the next layer. The rationale behind this method is to allow each convolution module to generate its own relevant features.

    \item \textit{Late fusion} methods rely on using only the predictions (scores or probability vector) of different sensor-specific models. As they allow to use any type of models (both Machine Learning and Deep learning), these methods are the most flexible category. 
    \begin{itemize} 
        \item \textit{Probability fusion}: Each network produces a probability vector (after the softmax), indicating the prediction of the network. With the probability fusion, we simply compute an average of probabilities. 

        \item \textit{Scores fusion}: To merge the scores, we extract the vector before the softmax. This vector of scores (or logits) indicates whether the sample is likely to belong to each class. We compute the average, for each class, of the score each network assigns to each class, before using a softmax to obtain a single output probability. 
        
        \item \textit{Weighted fusions}: With the two previous modes (probabilities and scores fusion), the average was unweigthed, which means a 'bad' sensor is given as much importance as a relevant one. To avoid it, we try letting the network learn the weight to assign to each sensor. With both methods, we compute a weighted sum of the predictions of the sensor-specific models.
    \end{itemize}
\end{itemize}

We also included several algorithms from the literature whose motivations apply to our work. These works are the Bottleneck filters  \cite{wang_deep_2019}, Attention \cite{xu_show_2016}, Selective fusion \cite{chen_selective_2019}, Learn to combine modalities in multimodal deep learning \cite{liu_learn_2018}, and Gradient Blend \cite{wang_what_2019}. In total, we evaluate 13 fusion modes. There is no need for a baseline in this case, because we choose to evaluate sensor choice and preprocessing method  on single-sensor data.

\section{Results} \label{sec:experiments_results}

\subsection{Evaluation of unimodal sensors}\label{subsec:results_per_sensor}

\begin{table*}[t]
\centering
\renewcommand{\arraystretch}{1.5}
\caption{The validation F1-score per signal. The highest result is in bold}
\scriptsize \renewcommand\theadfont{}
\begin{tabular}{|c|c|c|c|c|c|c|c|}
    \hline
    sensor & Acc & LAcc & Gra & Gyr & Mag & Ori & Pressure \\ \hline
    $x$    & $87.24 \pm 0.53$ & $83.97 \pm 0.80$ & $85.19 \pm 0.26$ & $81.31 \pm 0.57$ & $71.14 \pm 0.67$ & $73.82 \pm 1.24$ &   \\ \cline{1-7}
    $y$    & $87.22 \pm 0.72$ & $86.22 \pm 0.84$ & $84.44 \pm 0.22$ & $81.05 \pm 1.25$ & $73.63 \pm 0.82$ & $74.37 \pm 1.25$ &   \\ \cline{1-7}
    $z$    & $84.18 \pm 0.77$ & $85.36 \pm 0.69$ & $83.34 \pm 0.54$ & $79.32 \pm 1.32$ & $73.17 \pm 1.03$ & $75.46 \pm 0.51$ & $76.35 \pm 0.67$ \\ \cline{1-7}
    $norm$ & $\mathbf{89.14 \pm 0.65}$ & $81.01 \pm 0.50$ & $47.67 \pm 3.27$ & $76.52 \pm 0.68$ & $66.81 \pm 0.47$ & $42.05 \pm 0.97$ &   \\ \cline{1-7}
    $w$    &                           \multicolumn{5}{l|}{}                                              & $78.54 \pm 1.07$ &   \\ \hline
\end{tabular}
\label{tab:results_per_sensor}
\normalsize \renewcommand\theadfont{}
\end{table*}

Table \ref{tab:results_per_sensor} shows the norm of the accelerometer is the single best signal available. If the accelerometer is the best sensor, the linear acceleration and gravity follow closely. Also, the pressure signal is surprisingly effective at distinguishing between transport modes. 

For the accelerometer, the norm of the acceleration vector has a better performance than any of its individual axes, which shows the orientation-independent signal is better than its components. For the norm of the magnetometer, however, the result is the opposite, as the norm of the magnetometer is worse than any of its individual axes. 

One other surprisingly high performance is the norm of gravity and norm of the orientation vector. In theory, these two signals are constant (equal to 1 for the norm of the orientation vector, and $9.81 m.s^{-2}$ for the gravity), but in practice, these signals come from imperfect computations which yield indications about the dynamics of the sensor.

\subsection{Evaluation of preprocessing methods}\label{subsec:results_preprocessing}

\begin{table*}[t]
\centering
\caption{The validation F1-score (\%) per preprocessing method. For each signal, the highest result is in bold, and the second highest result is underlined}
\scriptsize \renewcommand\theadfont{}
\renewcommand{\arraystretch}{1.5}
\begin{tabular}{|c|c|c|c|c|c|c|c|}
    \hline
    mode        &  \thead{interpolation \\(frequency axis)} & power scale & \thead{size \\(T, F)}  & $|Acc|$            & $Gyr_y$            & $|Mag|$               & $Ori_w$  \\ \hline
    temporal    &                                           &             &           6000         & $ 70.20 \pm 1.63 $ & $ 64.71 \pm 2.74 $ & $ 7.49 \pm 10.32 $& $ 39.65 \pm 2.26 $ \\ \hline
    FFT         &                                           &             &           6000         & $ 80.57 \pm 1.30 $ & $ 74.30 \pm 0.72 $ & $ 55.99 \pm 1.53 $ & $ 64.27 \pm 1.57 $ \\ \hline
    spectrogram &                   none                    &    linear   &         550, 250       & $ 84.55 \pm 1.03 $ & $ 71.81 \pm 0.64 $ & $ 63.64 \pm 0.96 $ & $  2.29 \pm 0.00 $ \\ \hline
    spectrogram &                   none                    &     log     &         550, 250       & $ \underline{87.88 \pm 0.68 }$ & $ \underline{79.89 \pm 1.30} $ & $ \underline{64.81 \pm 0.85}$ & $ 43.35 \pm 33.56$ \\ \hline
    spectrogram &                  linear                   &    linear   &          48, 48        & $ 81.98 \pm 0.75 $ & $ 58.42 \pm 1.31 $ & $ 45.97 \pm 1.88 $ & $  2.29 \pm 0.00$ \\ \hline
    spectrogram &                  linear                   &     log     &          48, 48        & $ 86.33 \pm 1.00 $ & $ 77.06 \pm 1.32 $ & $ 56.53 \pm 0.68 $ & $ \underline{75.03 \pm 2.08}$ \\ \hline
    spectrogram &                log-freq              &    linear   &          48, 48        & $ 84.46 \pm 0.63 $ & $ 69.19 \pm 0.89 $ & $ 53.36 \pm 1.41 $ & $  2.29 \pm 0.00$ \\ \hline
    spectrogram &                log-freq              &     log     &          48, 48        & $\mathbf{ 88.83 \pm 0.71 }$ & $\mathbf{ 82.64 \pm 0.68 }$ & $\mathbf{67.36 \pm 0.49 }$ & $\mathbf{ 78.39 \pm 1.79 }$  \\ \hline
\end{tabular}
\normalsize \renewcommand\theadfont{}
\label{tab:results_preprocessing}
\end{table*}

Table \ref{tab:results_preprocessing} gives the results the evaluation of each preprocessing method. Note that the results in table \ref{tab:results_preprocessing} are similar to those obtained by Richoz \textit{et al.} \cite{richoz_transportation_2020} with a different neural network architecture: with three sensors (accelerometer, magnetometer, gyrometer), they obtained a F1-score of 79.4 \%. For the record, we reproduced their approach (frequency concatenation of FFT segments), with our setting (architecture from \cite{ito_application_2018}, Fourier transforms of 60-seconds long segments instead of 5s), and obtained a validation F1-score of $81.44 \pm 1.06 \%$.

Switching to the norm of the FFT is strictly better than using raw, temporal representations. This difference might be due to the fact that the power spectrum better separates patterns from noise (see fig. \ref{fig:avg_spectrum}).

But using spectrograms seems better, in most cases, than using a power spectrum, or even a temporal segment. One notable exception, however, is the raw, full-size spectrogram, with the orientation vector. This method has an impressive standard deviation, because two of the five initializations were failure cases that did not learn efficiently and had a F1-score of 2.8\% (which is the score of a classifier that predicts the most occurring mode). The others had a F1-score of $76.4 \pm 1.6\%$, which is closer to what one could expect given the results of the other sensors. We could not find any explanations to this behaviour, which is all the more surprising as neural networks are usually stable throughout different initializations (as our other experiment show).

Using the log of the power is strictly better than the raw power for the accelerometer, gyrometer, magnetometer: the average gain obtained by switching from the raw power to the log power is $7.66$ percentage points.

Using a logarithmic interpolation for the frequency axis allows to effectively reduce the size of the data without altering the signal as a linear interpolation does. This might be due to the fact that interpolating linearly a (550, 250) spectrogram into a (48, 48) matrix erases the difference between the fundamental frequency of the Walk and Bike segments (fig. \ref{fig:avg_spectrum}). A log scale preserves the distinction between these modes by giving more room to the lower frequencies.

\subsection{A benchmark of fusion modes}\label{subsec:results_fusion_modes}

   \begin{table*}[t]
    \centering
    \renewcommand{\arraystretch}{1.5}
    \caption{The mean and standard deviation of the validation F1-score of each method, for different fusion modes. For each sensor combination, we display the best result in bold and underline  the results that are less than two standard deviations away from the best}
    \tiny \renewcommand\theadfont{}
    \begin{tabular}{|c|c|c|c|c|c|c|c|c|c|c|c|c|c|}
        \hline
           & \multicolumn{4}{c|}{early fusion} & \multicolumn{3}{c|}{intermediate fusion} & \multicolumn{6}{c|}{late fusion} \\ \cline{2-14}
         method  & \thead{time\\concat.} & \thead{freq\\concat.} & \thead{depth\\concat.} & \thead{Bottleneck\\filters} & features & \thead{Selective\\Fusion} & attention &  probas & scores & \thead{Weighted\\probas} & \thead{Weighted\\scores} & \thead{Gradient\\Blend} & \thead{Learn to\\combine}  \\  \hline

        $|Acc|, Gyr_y$                            & \thead {$\underline{90.88}$ \\ $ \underline{\pm0.57}$} & \thead {$\underline{90.46}$ \\ $\underline{\pm1.08}$} & \thead {$\underline{90.57}$ \\ $\underline{\pm0.94}$} & \thead {$\underline{88.70}$ \\ $\underline{\pm1.30}$} & \thead {$\underline{90.83}$ \\ $\underline{\pm1.54}$} & \thead {$\underline{90.01}$ \\ $\underline{\pm0.41}$} & \thead {$84.11$ \\ $ \pm1.30$} & \thead {$\underline{90.26}$ \\ $\underline{\pm0.56}$} & \thead {$\mathbf{90.95}$ \\ $\mathbf{\pm0.37}$} & \thead {$88.85$ \\ $ \pm0.54$} & \thead {$\underline{90.89}$ \\ $\underline{\pm1.13}$} & \thead {$89.18$ \\ $ \pm1.10$} & \thead {$90.10$ \\ $ \pm0.66$} \\ \hline 
        
        $|Acc|, |Mag|$                            & \thead {$90.78$ \\ $ \pm0.66$} & \thead {$\underline{91.37}$ \\ $\underline{\pm0.49}$} & \thead {$\underline{91.83}$ \\ $ \underline{\pm0.35}$} & \thead {$85.83$ \\ $ \pm4.45$} & \thead {$\underline{91.74}$ \\ $\underline{\pm0.46}$} & \thead {$\underline{90.62}$ \\ $ \underline{\pm1.03}$} & \thead {$86.67$ \\ $ \pm1.01$} & \thead {$\underline{91.83}$ \\ $\underline{\pm0.73}$} & \thead {$\underline{91.72}$ \\ $ \underline{\pm0.41}$} & \thead {$88.04$ \\ $ \pm0.93$} & \thead {$\mathbf{92.17}$ \\ $ \mathbf{\pm0.59}$} & \thead {$89.53$ \\ $ \pm0.64$} & \thead {$87.37$ \\ $ \pm0.65$} \\ \hline

        \thead{$|Acc|, Gyr_y,$\\$ |Mag|$ }        & \thead {$\underline{91.36}$ \\ $ \underline{\pm0.74}$} & \thead {$\underline{92.13}$ \\ $\underline{\pm0.90}$} & \thead {$\underline{91.91}$ \\ $\underline{\pm0.88}$} & \thead {$87.01$ \\ $ \pm2.16$} & \thead {$\underline{91.87}$ \\ $\underline{\pm0.64}$} & \thead {$\underline{92.39}$ \\ $ \underline{\pm0.87}$} & \thead {$87.85$ \\ $ \pm1.05$} & \thead {$\underline{92.33}$ \\ $\underline{\pm0.61}$} & \thead {$\underline{92.55}$ \\ $\underline{\pm1.08}$} & \thead {$89.40$ \\ $ \pm0.55$} & \thead {$\mathbf{92.98}$ \\ $\mathbf{ \pm0.37}$} & \thead {$89.47$ \\ $ \pm0.92$} & \thead {$89.56$ \\ $ \pm0.98$} \\ \hline 
        
        \thead{$|Acc|, Gyr_y,$\\$ |Mag|, Ori_w$ } & \thead {$\underline{92.32}$ \\ $ \underline{\pm1.18}$} & \thead {$\underline{92.30}$ \\ $\underline{\pm0.54}$} & \thead {$\underline{91.23}$ \\ $ \underline{\pm1.25}$} & \thead {$84.59$ \\ $ \pm5.52$} & \thead {$\underline{92.51}$ \\ $\underline{\pm1.01}$} & \thead {$\underline{92.93}$ \\ $ \underline{\pm0.60}$} & \thead {$87.56$ \\ $ \pm0.62$} & \thead {$\underline{92.43}$ \\ $\underline{\pm0.40}$} & \thead {$\underline{92.83}$ \\ $\underline{\pm0.19}$} & \thead {$89.43$ \\ $ \pm0.49$} & \thead {$\mathbf{93.01}$ \\ $ \mathbf{\pm0.36}$} & \thead {$89.36$ \\ $ \pm1.24$} & \thead {$\underline{91.41}$ \\ $\underline{\pm1.11}$} \\ \hline 
    
    \end{tabular}
    \normalsize \renewcommand\theadfont{}
    \label{tab:results_fusion_modes}
    \end{table*}

Table \ref{tab:results_fusion_modes} gives the results of each fusion method, applied to four sensor combinations. We notice that, given the standard deviation of the experiments, most fusion methods have statistically similar performance. In fact, data fusion may be the least determinant choice we evaluated (see table \ref{tab:influence_parameters}). 
As most fusion methods have the same performance, we recommend using the most simple ones.

\begin{table}[t]
    \centering
    \caption{A summary of the influence of the parameters in tables  \ref{tab:results_per_sensor}, \ref{tab:results_preprocessing}, and \ref{tab:results_fusion_modes}. We purposely excluded the scenarios where the network did not learn anything (defined as F1-score $< 10 \%$)}
    \renewcommand{\arraystretch}{1.2}
    \scriptsize \renewcommand\theadfont{}
    \begin{tabular}{|c|c|c|}
        \hline
        \multicolumn{3}{|c|}{Average gain by switching:} \\ \hline
        from &  to  & \\  \hline
        $Gyr_y$ & $|Acc|$ & $+12.55$ \\  \hline
        $|Mag|$ & $|Acc|$ & $+27.06$ \\  \hline
        $Ori_w$ & $|Acc|$ & $+20.06$ \\  \hline
        spectrogram raw power & spectrogram log power & $+8.66$ \\  \hline
        $250 \times 550$ spectrogram (log power) & $48 \times 48$ spectrogram (linear interp., log power) & $-4.22$ \\  \hline
        $250 \times 550$ spectrogram (log power) & $48 \times 48$ spectrogram (log interp., log power) &  $+2.08$ \\  \hline
        raw (temporal) data & $48 \times 48$ spectrogram (log interp., log power) & $ +25.10 $ \\  \hline
        Median fusion method & Best fusion method & $+1.12$ \\  \hline
        
    \end{tabular}
    \label{tab:influence_parameters}
    \normalsize \renewcommand\theadfont{}
\end{table}

\subsection{Evaluations on the test set}\label{subsec:test_results}

In order to compare our approach against the state of the art, we select the best fusion method and sensor combination in table \ref{tab:results_fusion_modes} (that is, a weighted score fusion of the four sensors), and train it 5 times on the union of train and validation sets, before evaluating the F1 score on the held-out test set of the challenge. If the F1-score is still significantly lower than it was on the validation set, the loss is acceptable ($89.96 \pm 0.07 \%$ and $93.01 \pm 0.36$, respectively). However, the best model is barely superior to a single-sensor baseline ($87.44 \pm 0.98 \%$ on test, $89.14 \pm 0.65$ on val), which consists in using the accelerometer alone. In practice, the use of additional sensors (and the energy consumption that comes with it) might not justify the performance gain.

\section{Conclusion}
We studied the application of Convolutional Neural Network on a Transport Mode Detection problem. By fixing all but one choice in each of our experiments, we could evaluate ech of the choices a practitioner can make. We found the more important choices to make are the sensor choice (the accelerometer being the best) and the preprocessing method (the use of spectrogram, with a logarithm axis for the frequencies), However, after evaluating 13 different data fusion methods, we found that no fusion method significantly outperforms the others. Future work might consist in adapting this benchmark to the newer datasets SHL 2019 and 2020, and the new research problems that come with them. Alternatively, we could extend our work to start from the approaches of \cite{gjoreski_applying_2018,gjoreski_classical_2020}, who used noticeably more complex methods to get better results on the challenge.

\bibliography{references}

\end{document}